\documentclass[square,authoryear,colon]{article}

\usepackage{iclr2017_arXiv,times}
\usepackage{hyperref}
\usepackage{url}
\usepackage{graphicx}
\usepackage{epstopdf}
\usepackage{algorithm2e}
\usepackage{booktabs}

\usepackage{bm}
\usepackage{amsmath}

\usepackage{xcolor}

\SetCommentSty{mycommfont}

\title{Place Recognition:An Overview of Vision Perspective}

\author{
  Zhiqiang Zeng (lbxzzq@163.com)\\
  Jian Zhang (jzhang@zisu.edu.cn)\\
  Xiaodong Wang (xdwangjsj@xmut.edu.cn)\\
  Yuming Chen (ymchen@xmut.edu.cn)\\
  Chaoyang Zhu (15051748@hdu.edu.cn) \\
  }

\begin{document}

\maketitle

\begin{abstract}

    Place recognition is one of the most fundamental topics in computer vision and robotics communities, where the task is to accurately and efficiently recognize the location of a given query image. Despite years of wisdom accumulated in this field, place recognition still remains an open problem due to the various ways in which the appearance of real-world places may differ. This paper presents an overview of the place recognition literature. Since condition invariant and viewpoint invariant features are essential factors to long-term robust visual place recognition system, We start with traditional image description methodology developed in the past, which exploit techniques from image retrieval field. Recently, the rapid advances of related fields such as object detection and image classification have inspired a new technique to improve visual place recognition system, i.e., convolutional neural networks (CNNs). Thus we then introduce recent progress of visual place recognition system based on CNNs to automatically learn better image representations for places. Eventually, we close with discussions and future work of place recognition.

\end{abstract}

\section{Introduction}\label{introduction}

    Place recognition has attracted a significant amount of attention in computer vision and robotics communities, as evidenced by related citations and a number of workshops dedicated to improve long-term robot navigation and autonomy. It has a number of applications ranging from autonomous driving, robot navigation to augmented reality, geo-localizing archival imagery.

\begin{figure}[htb]
\begin{center}

        \includegraphics[width=13cm]{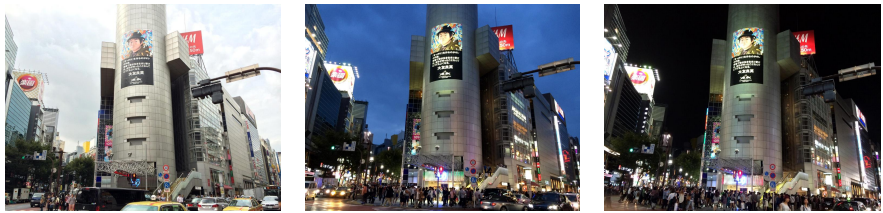}\label{illumination}
        \caption{TokyoTimeMachine dataset \cite{NetVLAD} : images from the same place with different illumination conditions : day, sunset, night.}

\end{center}
\end{figure}

\begin{figure}[htb]
\begin{center}

        \includegraphics[width=14cm]{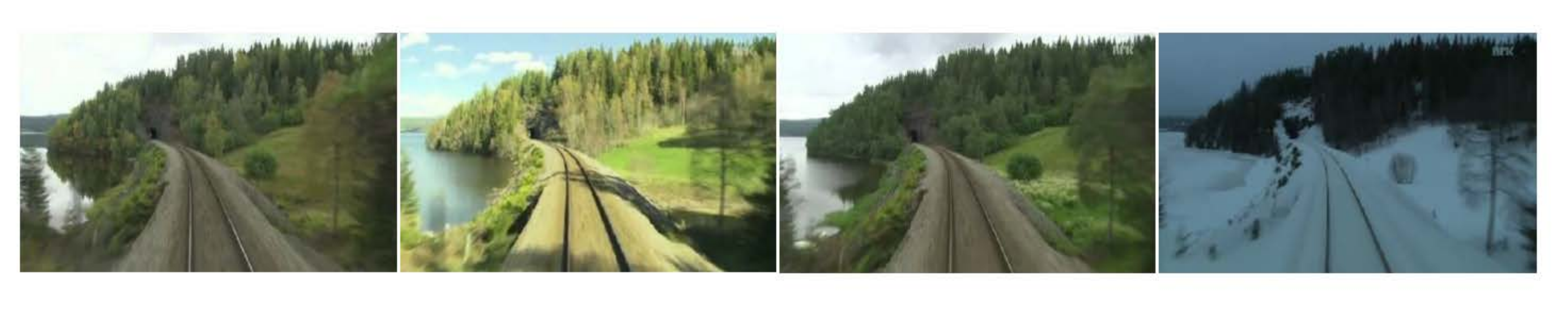}\label{weather}
        \caption{Frames extracted from the Nordland dataset \cite{Arewethereyet} that belong to the same place in spring, summer, fall, winter.}

\end{center}
\end{figure}

    The process of identifying the location of a given image by querying the locations of images belonging to the same place in a large geotagged database, usually known as place recognition, is still an open problem. One major characteristic that separates place recognition from other visual recognition tasks is that place recognition has to solve condition invariant recognition to a degree that many other fields haven't. How can we robustly identify the same real-world place undergoing major changes in appearance, e.g., illumination variation (Figure ~\ref{illumination}), change of seasons (Figure ~\ref{weather}) or weather, structural modifications over time, and viewpoint change. To be clear, above changes in appearance are summarized as conditional variations, but excludes the viewpoint change. Moreover, how can we distinguish true images from similarly looking images without supervision? Since collecting geotagged datasets is time-consuming and labor intensive, and situations like indoor places do not have GPS information necessarily. Place recognition task has been traditionally cast as an image retrieval task \cite{Yu2015a} where image representations for places are essential. The fundamental scientific question is that what is the appropriate representation of a place that is informative enough to recognize real-world places, yet compact enough to satisfy the real-time processing requirement on a terminal, such as mobile phone and a robot.

    At early stages, place recognition was dominated by sophisticated local-invariant feature extractors such as SIFT \cite{SIFT} and SURF \cite{SURF}, hand-crafted global image descriptors such as GIST \cite{Oliva2001,Oliva2006}, and bag-of-visual-words \cite{BoW1,BoW2} approach. These traditional feature extraction techniques have led to a great step towards the ultimate outcome.

    Recent years have witnessed a prosperous advancement of visual content recognition using a powerful image representations extractor - Convolutional Neural Networks (CNNs) \cite{ImageNet,vgg,Yu2017,Yu2017a}, which sets state-of-the-art performance on many category-level recognition tasks such as object classification \cite{ImageNet,vgg,googlenet}, scene recognition \cite{places205,scenerecognition,Yu2015}, image classification \cite{Yu2012}. Principle ideas of CNNs can date back to 1980s, and the major two reasons why CNNs are so successful in computer vision are the advances of GPU-based computation power and data volume respectively. Recent studies show that general features extracted by CNNs can be transferable \cite{CNN-off-the-shelf} and generalized well to other visual recognition tasks. Semantic gap is a well-known problem in place recognition, where different semantics of places may share common low-level features extracted by SIFT, e.g., colors, textures. Convolutional Neural Networks may bridge this semantic gap by treating an image as an high-level feature vector, extracted through deep stacked layers.
		
    This paper provides an overview of both traditional and deep learning based descriptive techniques widely applied to place recognition task, which by no means is exhaustive. The remainder of this paper is organized as follows : section 1 gives an introduction of place recognition literature. Section 2 talks about local and global image descriptors widely applied to place recognition. Section 3 presents a brief view of convolutional neural networks and corresponding techniques used in place recognition. Section 4 discusses the future work in place recognition.

\section{Traditional Image Descriptors}

\subsection{Local Image Descriptors}\label{localimagedescriptors}

    Local feature descriptors such as SIFT \cite{SIFT} and SURF \cite{SURF} have been widely applied to visual localization and place recognition tasks. They achieve an outstanding performance on viewpoint invariance. SIFT and SURF describes the local appearance of individual patches or key points within an image, for an image, $\mathbf{X}=\mathbf{[\mathbf{x}_{1},\mathbf{x}_{2},...,\mathbf{x}_{N}]}^{T}$ represents the set of local invariant features, in which $\mathbf{x}_{i} \in \mathbf{R}^{d}$ is the $\mathbf{i}^{th}$ local feature. The length of $\mathbf{X}$ depends on the number of key points of the image. These local invariant features are then usually aggregated into a compact single vector for the entire image, using techniques such as bag-of-visual-words \cite{BoW1,BoW2}, VLAD \cite{VLAD} and Fisher kernel \cite{FV1,FV2}.

    Since local feature extraction consists of two phases : detection and description, a number of variations and extensions of techniques for the two phases are developed. For example, \cite{constant} used FAST \cite{fast} to detect the interested patches of an image, which were then described by SIFT. \cite{experience} used FAST as well during detection phase, whereas they use BRIEF \cite{brief} to describe the key points instead of SIFT.

    State-of-the-art visual simultaneous localization and mapping (SLAM) systems such as FAB-MAP \cite{FAB-MAP,FAB-MAP2.0} used bag-of-visual-words to construct the final image representation. Each image may contain hundreds of local features, which is impractical in large-scale and real-time processing place recognition task, moreover, they require an enormous amount of memory to store the high dimensional features. The bag-of-visual-words mimics the bag-of-words technique used in efficient text retrieval field. It typically needs to form a codebook $\mathbf{C}=\mathbf{[\mathbf{c}_{1},\mathbf{c}_{2},...,\mathbf{c}_{k}]}$ with $\mathbf{K}$ visual words, each visual word $\mathbf{c}_{i} \in \mathbf{R}^{d}$ is a centroid of a cluster, usually gained by k-means \cite{k-means}. Then each local invariant feature $\mathbf{x}_{i}$ is assigned to its nearest cluster centroid, i.e., the visual word. A histogram vector of $\mathbf{k}$ dimension containing the frequency of each visual word being assigned can be formed this way. The bag-of-visual-words model and local image descriptors generally ignore the geometric structure of the image, that is, different orders of local invariant features in $\mathbf{X}$ will not impact on the histogram vector, thus the resulting image representation is viewpoint invariant. There are several variations on how to normalize the histogram vector, a common choice is $\mathbf{L}_{2}$ normalization. Components of the vector are then weighted by inverse document frequency (idf). However, the codebook is dataset dependent and needs to be retrained if a robot moves into a new region it has never seen before. The lack of structural information of the image can also weaken the performance of the model.

    Fisher Kernel proposed by \cite{FV1,FV2} is a powerful tool in pattern classification combining the strengths of generative models and discriminative classifiers. It defines a generative probability model $\mathbf{p}$, which is the probability density function with parameter $\mathbf{\lambda}$. Then one can characterize the set of local invariant features $\mathbf{X}=\mathbf{[\mathbf{x}_{1},\mathbf{x}_{2},...,\mathbf{x}_{N}]}^{T}$ with the following gradient vector :

\begin{equation}\label{FisherVector}
\mathbf{\nabla}_{\lambda}\log_{}\mathbf{p(\mathbf{X}|\mathbf{\lambda})}
\end{equation}

    where the gradient of the log-likelihood describes the direction in which parameters should be modified to best fit the observed data intuitively. It transforms a variable length sample $\mathbf{X}$ into a fixed length vector whose size is only dependent on the number of parameters in the model.

    \cite{FV2} applied Fisher kernel in the context of image classification with a Gaussian Mixture Model to model the visual words. In comparison with the bag-of-visual-words representation, they obtain a $\mathbf{(2d+1)*k-1}$ dimensional image representation of a local invariant feature set, while $\mathbf{k}$ dimensional image representation using bag-of-visual-words. Thus, fisher kernel can provide richer information under the circumstance that their size of codebook is equal, or, fewer visual words are required by this more sophisticated representation.

    \cite{VLAD} propose a new local aggregation method called Vectors of locally aggregated descriptors (VLAD), which becomes state-of-the-art technique compared to bag-of-visual-words and fisher vector. The final representation is computed as follows:

\begin{equation}\label{VLAD}
\mathbf{v}_{i,j}=\mathbf{\sum_{\mathbf{x} such that NN(\mathbf{x})=\mathbf{c}_{i}}}\mathbf{x}_{j}-\mathbf{c}_{i,j}
\end{equation}

     The VLAD vector is represented by $\mathbf{v}_{i,j}$ where the indices $i=1...k$ and $j=1...d$ respectively index the $\mathbf{i}^{th}$ visual word and the $\mathbf{j}^{th}$ component of local invariant feature. The vector is subsequently $\mathbf{L}_{2}$ normalized. We can see that the final representation stores the sum of all the residuals between local invariant features and its nearest visual word. One excellent property of VLAD vector is that it's relatively sparse and very structured, \cite{VLAD} show that a principle component analysis is likely to capture this structure for dimensionality reduction without much degradation of representation. They obtain a comparable search quality to bag-of-visual-words and Fisher kernel with at least an order of magnitude less memory.

\subsection{Global Image Descriptors}\label{globalimagedescriptors}

    The key difference between local place descriptors and global place descriptors is the presence of detection phase. One can easily figure out that local place descriptors turns into global descriptors by predefining the key points as the whole image. WI-SURF \cite{wi-surf} used whole-image descriptors based on SURF features and BRIEF-GIST \cite{brief-gist} used BRIEF \cite{brief} features in a similar whole-image fashion.

    A representative global descriptor is GIST \cite{Oliva2001,Oliva2006}. It has been shown to suitably model semantically meaningful and visually similar scenes in a very compact vector. The amount of perceptual and semantic information that observers comprehend within a glance (around 200ms) refers to the gist of the scene, termed Spatial Envelope properties, it encodes the dominant spatial layout information of the image. GIST uses Gabor filters at different orientations and different frequencies to extract information from the image. The results are averaged to generate a compact vector that represents the "gist" of a scene. \cite{experimentsinprusinggistpanoramas} applied GIST into large-scale omnidirectional imagery and obtains a nice segmentation of the search space into clusters, e.g., tall buildings, streets, open areas, mountains. \cite{biologically} followed a biological strategy which first computes the gist of a scene to produce a coarse localization hypothesis, then refine it by locating salient landmark points in the scene.

    The performance of techniques described in section 2 mainly depends on the size of codebook $\mathbf{C}$, if too small, the codebook will not characterize the dataset well, on the contrast, if the size is too large, it will require huge computational resources plus time. While global image descriptors have their own disadvantages, they usually assume that images are taken from a same viewpoint.

\section{Convolutional Neural Networks}\label{CNNs}

\begin{figure}[!htb]
\centering
\includegraphics[width=0.8\linewidth]{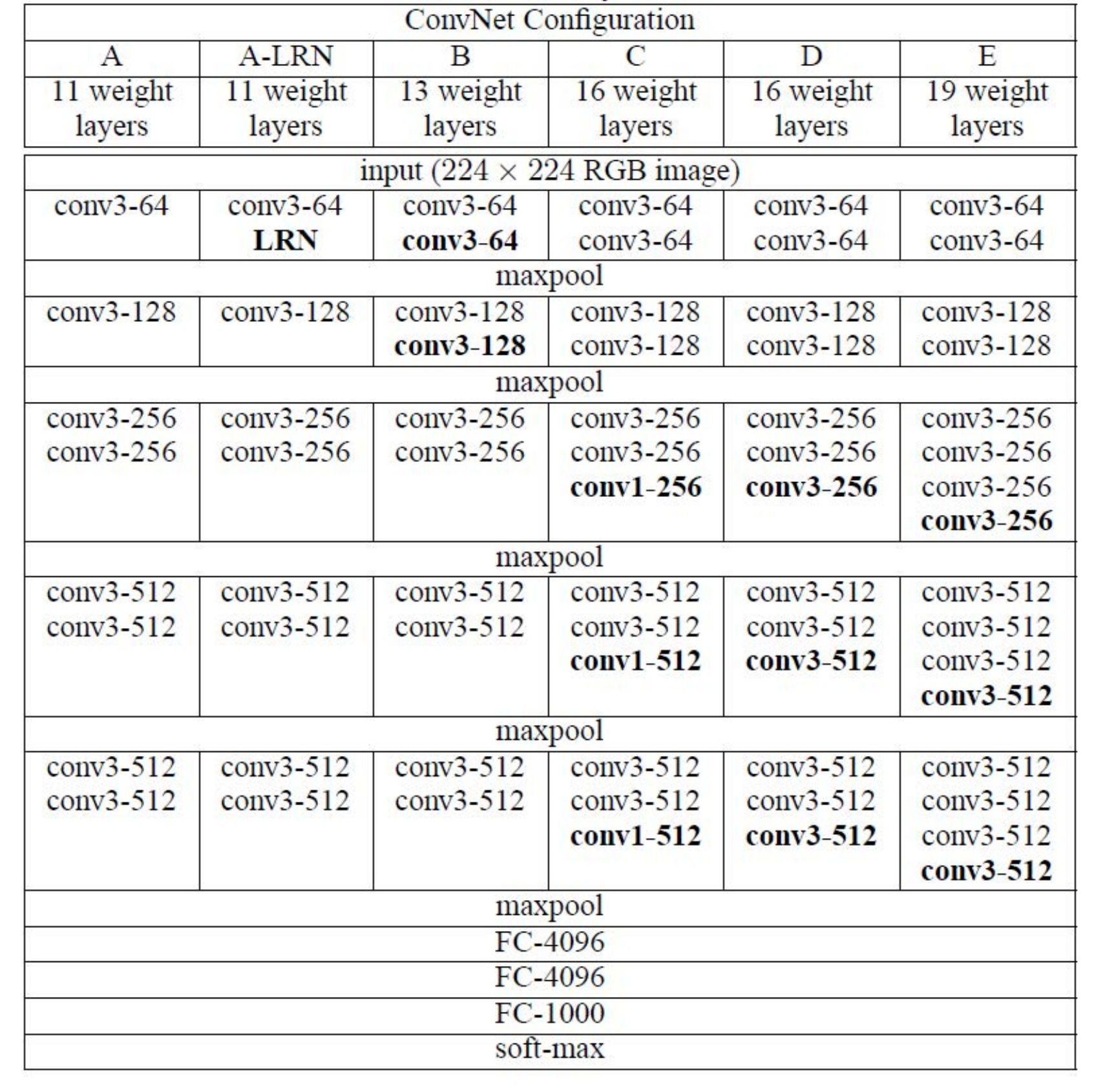}
\caption{VGG-16 configuration (shown in columns). The depth of the configuration increases from the left(A) to the right(E), as more layers are added(the added layers are shown in bold). The convolutional layer parameters are denoted as "conv(receptive field size)-(number of channels)". The ReLU activation function is omitted for simplicity.}\label{vgg16}
\end{figure}

    Recently Convolutional Neural Networks achieve state-of-the-art performance on various classification and recognition tasks, e.g., handwriting digits recognition, object classification \cite{ImageNet,vgg,googlenet}, scene recognition \cite{places205,scenerecognition}. Features extracted from convolutional neural networks trained on very large datasets significantly outperforms SIFT on a variety of vision tasks \cite{ImageNet,Fischer2014}. The core idea behind CNNs is the ability to automatically learn high-level features trained on a significant amount of data through deep stacked layers in an end-to-end manner. It works as a function $\mathbf{f(\cdot)}$ that takes some inputs such as images and output the image representations characterized by a vector. A common CNN model for fine-tuning is vgg-16 \cite{vgg}, the architecture can be seen from Figure ~\ref{vgg16}. For an intuitive understanding of what the CNN model learns in each layer, please check \cite{Convnet,NetVLAD} for heatmap graphical explanation.

    \cite{CNNPR} was the first work to exploit CNN in place recognition system as a feature extractor. They used a pre-trained CNN called Overfeat \cite{OverFeat}, which is originally proposed for the ImageNet Large Scale Visual Recognition Challenge 2013 (ILSVRC2013) and proved that advantages of deep learning can shift to place recognition task. \cite{Fischer2014} provides an investigation on the performance of CNN and SIFT on a descriptor matching benchmark. \cite{Convnet} comprehensively evaluates and compares the utility and viewpoint-invariant properties of CNNs.	 They've shown that features extracted from middle layers of CNNs have a good robustness against conditional changes, including illumination change, seasonal and weather changes, while features extracted from top layers are more robust to viewpoint change. Features learnt by CNNs are proved to be versatile and transferable, i.e., even though they were trained on a specific target task, they can be successfully deployed for other problems and often outperform traditional hand-engineered features \cite{CNN-off-the-shelf}. However, their usage as black-box descriptors in place recognition has so far yielded limited improvements. For example, visual cues that are relevant for object classification may not benefit the place recognition task. \cite{NetVLAD} design a new CNN architecture based on vgg16 and the VLAD representation, they remove the all the fully connected layers and plug the VLAD layers into it by making it differentiable. The loss function used in this paper is triplet loss, which can be seen in many other recognition tasks. \cite{places205} gathered a huge scene-centric dataset called "Places" containing more than 7 million images from 476 scene categories. However, scene recognition is fundamentally different from place recognition. \cite{DLFSVP} creates for the first time a large-scale place-centric dataset called SPED containing over 2.5 million images.

    Though CNNs has the power to extract high-level features, we are far away from making full use of it. How to gather a sufficient amount of data for place recognition, how to train a CNN model in an end-to-end manner to automatically choose the optimal features to represent the image, are still underlying problems to be solved.

\section{Discussion and Future Work}

    Place recognition has made great advances in the last few decades, e.g., the principles of how animals recognize and remember places and relationship between places in a neuroscience perspective \cite{survey}, a new way of describing places using convolutional neural networks, a number of datasets specifically for places are put forward. However, we are still a long way from a robust long-term visual place recognition system which can be well applied to a variety of scenarios of real-world places. Hence, we highlight several promising avenues of ongoing and future research that are leading us closer to this ultimate outcome.

    Place recognition is becoming a hot research field and benefiting from related ongoing works in other fields, especially the enormous successes achieved in computer vision through deep learning technique, e.g., image classification, object detection, scene recognition. While features extracted from pre-trained CNNs on other vision tasks are shown to be transferable and versatile, they've so far yielded unsatisfactory performance on place recognition task. There is a high possibility that we still don't fully exploit the potential of CNNs, we can improve the performance in two aspects. First, gather a sufficient amount of place-centric data covering various environments including illumination, weather, structure, season and viewpoint change. One alternative and reliable source is Google Street View Time Machine. If you train a CNN on a small-size dataset, the model usually works awful on other datasets since place recognition is dataset-dependent. And one needs to retrain it when a new dataset is fed into it. Since one of advantages of CNNs is to extract representative features through Big data, state-of-the-art performance can be improved. Second, an optimized CNN architecture for place recognition task. Real-world images from cameras are usually high resolution, whereas in many cases one needs to downscale the original images. For example, the input size of vgg-16 is 224*224. Moreover, an architecture that is well-suited for object detection may not fit well into place recognition task since their visual cues are different. And designing a good loss function is essential to features. Developments focused on the above two problems will further improve the robustness and performance of existing deep learning-based place recognition techniques.

    Place recognition system can also benefit from ongoing researches of object detection, scene classification \cite{Yu2013} and scene recognition \cite{Yu2015}. Semantic context from scene, interpreted as the "gist" of the scene, can help to partition the search space when comparing similarity between image representations, which ensures scalability and real-time processing towards real-world application. Note that different places may share a common semantic concept, which needs a furthermore and precise feature mapping procedure. Objects such as pedestrian and trees shoulsd be avoided, while objects like buildings and landmarks are important for long-term place recognition. Automatically determining and suppressing  features that leads to confusion to visual place recognition systems will also improve the place recognition performance.

\bibliographystyle{plainnat}
\bibliography{PRoverview}

\begin{thebibliography}{40}
\providecommand{\natexlab}[1]{#1}
\providecommand{\url}[1]{\texttt{#1}}
\expandafter\ifx\csname urlstyle\endcsname\relax
  \providecommand{\doi}[1]{doi: #1}\else
  \providecommand{\doi}{doi: \begingroup \urlstyle{rm}\Url}\fi

\bibitem[Arandjelovi\'c et~al.(2016)Arandjelovi\'c, Gronat, Torii, Pajdla, and
  Sivic]{NetVLAD}
R.~Arandjelovi\'c, P.~Gronat, A.~Torii, T.~Pajdla, and J.~Sivic.
\newblock {NetVLAD}: {CNN} architecture for weakly supervised place
  recognition.
\newblock In \emph{IEEE Conference on Computer Vision and Pattern Recognition},
  2016.

\bibitem[Badino et~al.(2012)Badino, Huber, and Kanade]{wi-surf}
Hern{\'a}n Badino, Daniel Huber, and Takeo Kanade.
\newblock Real-time topometric localization.
\newblock In \emph{Robotics and Automation (ICRA), 2012 IEEE International
  Conference on}, pages 1635--1642. IEEE, 2012.

\bibitem[Bay et~al.(2006)Bay, Tuytelaars, and Van~Gool]{SURF}
Herbert Bay, Tinne Tuytelaars, and Luc Van~Gool.
\newblock Surf: Speeded up robust features.
\newblock \emph{Computer vision--ECCV 2006}, pages 404--417, 2006.

\bibitem[Calonder et~al.(2012)Calonder, Lepetit, Ozuysal, Trzcinski, Strecha,
  and Fua]{brief}
Michael Calonder, Vincent Lepetit, Mustafa Ozuysal, Tomasz Trzcinski, Christoph
  Strecha, and Pascal Fua.
\newblock Brief: Computing a local binary descriptor very fast.
\newblock \emph{IEEE Transactions on Pattern Analysis and Machine
  Intelligence}, 34\penalty0 (7):\penalty0 1281--1298, 2012.

\bibitem[Chen et~al.(2014)Chen, Lam, Jacobson, and Milford]{CNNPR}
Zetao Chen, Obadiah Lam, Adam Jacobson, and Michael Milford.
\newblock Convolutional neural network-based place recognition.
\newblock \emph{CoRR}, abs/1411.1509, 2014.
\newblock URL \url{http://arxiv.org/abs/1411.1509}.

\bibitem[Chen et~al.(2017)Chen, Jacobson, S{\"{u}}nderhauf, Upcroft, Liu, Shen,
  Reid, and Milford]{DLFSVP}
Zetao Chen, Adam Jacobson, Niko S{\"{u}}nderhauf, Ben Upcroft, Lingqiao Liu,
  Chunhua Shen, Ian~D. Reid, and Michael Milford.
\newblock Deep learning features at scale for visual place recognition.
\newblock \emph{CoRR}, abs/1701.05105, 2017.
\newblock URL \url{http://arxiv.org/abs/1701.05105}.

\bibitem[Christian et~al.(2014)]{googlenet}
Szegedy Christian et~al.
\newblock Going deeper with convolutions. arxiv preprint.
\newblock \emph{arXiv preprint arXiv:1409.4842}, 2014.

\bibitem[Churchill and Newman(2013)]{experience}
Winston Churchill and Paul Newman.
\newblock Experience-based navigation for long-term localisation.
\newblock \emph{The International Journal of Robotics Research}, 32\penalty0
  (14):\penalty0 1645--1661, 2013.

\bibitem[Cummins and Newman(2008)]{FAB-MAP}
Mark Cummins and Paul Newman.
\newblock {FAB-MAP}: Probabilistic localization and mapping in the space of
  appearance.
\newblock \emph{The International Journal of Robotics Research}, 27\penalty0
  (6):\penalty0 647--665, 2008.
\newblock \doi{10.1177/0278364908090961}.

\bibitem[Cummins and Newman(2011)]{FAB-MAP2.0}
Mark Cummins and Paul Newman.
\newblock Appearance-only slam at large scale with fab-map 2.0.
\newblock \emph{The International Journal of Robotics Research}, 30\penalty0
  (9):\penalty0 1100--1123, 2011.

\bibitem[Fischer et~al.(2014)Fischer, Dosovitskiy, and Brox]{Fischer2014}
Philipp Fischer, Alexey Dosovitskiy, and Thomas Brox.
\newblock Descriptor matching with convolutional neural networks: a comparison
  to sift.
\newblock \emph{arXiv preprint arXiv:1405.5769}, 2014.

\bibitem[Jaakkola et~al.(1999)Jaakkola, Haussler, et~al.]{FV1}
Tommi~S Jaakkola, David Haussler, et~al.
\newblock Exploiting generative models in discriminative classifiers.
\newblock \emph{Advances in neural information processing systems}, pages
  487--493, 1999.

\bibitem[J{\'e}gou et~al.(2010)J{\'e}gou, Douze, Schmid, and P{\'e}rez]{VLAD}
Herv{\'e} J{\'e}gou, Matthijs Douze, Cordelia Schmid, and Patrick P{\'e}rez.
\newblock Aggregating local descriptors into a compact image representation.
\newblock In \emph{Computer Vision and Pattern Recognition (CVPR), 2010 IEEE
  Conference on}, pages 3304--3311. IEEE, 2010.

\bibitem[Kanungo et~al.(2002)Kanungo, Mount, Netanyahu, Piatko, Silverman, and
  Wu]{k-means}
Tapas Kanungo, David~M Mount, Nathan~S Netanyahu, Christine~D Piatko, Ruth
  Silverman, and Angela~Y Wu.
\newblock An efficient k-means clustering algorithm: Analysis and
  implementation.
\newblock \emph{IEEE transactions on pattern analysis and machine
  intelligence}, 24\penalty0 (7):\penalty0 881--892, 2002.

\bibitem[Krizhevsky et~al.(2012)Krizhevsky, Sutskever, and Hinton]{ImageNet}
Alex Krizhevsky, Ilya Sutskever, and Geoffrey~E Hinton.
\newblock Imagenet classification with deep convolutional neural networks.
\newblock In \emph{Advances in neural information processing systems}, pages
  1097--1105, 2012.

\bibitem[Lowe(2004)]{SIFT}
David~G Lowe.
\newblock Distinctive image features from scale-invariant keypoints.
\newblock \emph{International journal of computer vision}, 60\penalty0
  (2):\penalty0 91--110, 2004.

\bibitem[Lowry et~al.(2016)Lowry, S{\"u}nderhauf, Newman, Leonard, Cox, Corke,
  and Milford]{survey}
Stephanie Lowry, Niko S{\"u}nderhauf, Paul Newman, John~J Leonard, David Cox,
  Peter Corke, and Michael~J Milford.
\newblock Visual place recognition: A survey.
\newblock \emph{IEEE Transactions on Robotics}, 32\penalty0 (1):\penalty0
  1--19, 2016.

\bibitem[Mei et~al.(2009)Mei, Sibley, Cummins, Newman, and Reid]{constant}
Christopher Mei, Gabe Sibley, Mark Cummins, Paul~M Newman, and Ian~D Reid.
\newblock A constant-time efficient stereo slam system.
\newblock In \emph{BMVC}, pages 1--11, 2009.

\bibitem[Murillo and Kosecka(2009)]{experimentsinprusinggistpanoramas}
Ana~C Murillo and Jana Kosecka.
\newblock Experiments in place recognition using gist panoramas.
\newblock In \emph{Computer Vision Workshops (ICCV Workshops), 2009 IEEE 12th
  International Conference on}, pages 2196--2203. IEEE, 2009.

\bibitem[Oliva and Torralba(2001)]{Oliva2001}
Aude Oliva and Antonio Torralba.
\newblock Modeling the shape of the scene: A holistic representation of the
  spatial envelope.
\newblock \emph{International journal of computer vision}, 42\penalty0
  (3):\penalty0 145--175, 2001.

\bibitem[Oliva and Torralba(2006)]{Oliva2006}
Aude Oliva and Antonio Torralba.
\newblock Building the gist of a scene: The role of global image features in
  recognition.
\newblock \emph{Progress in brain research}, 155:\penalty0 23--36, 2006.

\bibitem[Perronnin and Dance(2007)]{FV2}
Florent Perronnin and Christopher Dance.
\newblock Fisher kernels on visual vocabularies for image categorization.
\newblock In \emph{Computer Vision and Pattern Recognition, 2007. CVPR'07. IEEE
  Conference on}, pages 1--8. IEEE, 2007.

\bibitem[Philbin et~al.(2007)Philbin, Chum, Isard, Sivic, and Zisserman]{BoW1}
James Philbin, Ondrej Chum, Michael Isard, Josef Sivic, and Andrew Zisserman.
\newblock Object retrieval with large vocabularies and fast spatial matching.
\newblock In \emph{Computer Vision and Pattern Recognition, 2007. CVPR'07. IEEE
  Conference on}, pages 1--8. IEEE, 2007.

\bibitem[Rosten and Drummond(2006)]{fast}
Edward Rosten and Tom Drummond.
\newblock Machine learning for high-speed corner detection.
\newblock \emph{Computer vision--ECCV 2006}, pages 430--443, 2006.

\bibitem[Sermanet et~al.(2013)Sermanet, Eigen, Zhang, Mathieu, Fergus, and
  LeCun]{OverFeat}
Pierre Sermanet, David Eigen, Xiang Zhang, Micha{\"e}l Mathieu, Rob Fergus, and
  Yann LeCun.
\newblock Overfeat: Integrated recognition, localization and detection using
  convolutional networks.
\newblock \emph{arXiv preprint arXiv:1312.6229}, 2013.

\bibitem[Sharif~Razavian et~al.(2014)Sharif~Razavian, Azizpour, Sullivan, and
  Carlsson]{CNN-off-the-shelf}
Ali Sharif~Razavian, Hossein Azizpour, Josephine Sullivan, and Stefan Carlsson.
\newblock Cnn features off-the-shelf: An astounding baseline for recognition.
\newblock In \emph{The IEEE Conference on Computer Vision and Pattern
  Recognition (CVPR) Workshops}, June 2014.

\bibitem[Siagian and Itti(2009)]{biologically}
Christian Siagian and Laurent Itti.
\newblock Biologically inspired mobile robot vision localization.
\newblock \emph{IEEE Transactions on Robotics}, 25\penalty0 (4):\penalty0
  861--873, 2009.

\bibitem[Simonyan and Zisserman(2014)]{vgg}
Karen Simonyan and Andrew Zisserman.
\newblock Very deep convolutional networks for large-scale image recognition.
\newblock \emph{arXiv preprint arXiv:1409.1556}, 2014.

\bibitem[Sivic et~al.(2003)Sivic, Zisserman, et~al.]{BoW2}
Josef Sivic, Andrew Zisserman, et~al.
\newblock Video google: A text retrieval approach to object matching in videos.
\newblock In \emph{iccv}, volume~2, pages 1470--1477, 2003.

\bibitem[S{\"u}nderhauf and Protzel(2011)]{brief-gist}
Niko S{\"u}nderhauf and Peter Protzel.
\newblock Brief-gist-closing the loop by simple means.
\newblock In \emph{Intelligent Robots and Systems (IROS), 2011 IEEE/RSJ
  International Conference on}, pages 1234--1241. IEEE, 2011.

\bibitem[S{\"u}nderhauf et~al.(2013)S{\"u}nderhauf, Neubert, and
  Protzel]{Arewethereyet}
Niko S{\"u}nderhauf, Peer Neubert, and Peter Protzel.
\newblock Are we there yet? challenging seqslam on a 3000 km journey across all
  four seasons.
\newblock In \emph{Proc. of Workshop on Long-Term Autonomy, IEEE International
  Conference on Robotics and Automation (ICRA)}, page 2013, 2013.

\bibitem[S{\"u}nderhauf et~al.(2015)S{\"u}nderhauf, Shirazi, Dayoub, Upcroft,
  and Milford]{Convnet}
Niko S{\"u}nderhauf, Sareh Shirazi, Feras Dayoub, Ben Upcroft, and Michael
  Milford.
\newblock On the performance of convnet features for place recognition.
\newblock In \emph{Intelligent Robots and Systems (IROS), 2015 IEEE/RSJ
  International Conference on}, pages 4297--4304. IEEE, 2015.

\bibitem[Yu et~al.(2012)Yu, Tao, and Wang]{Yu2012}
Jun Yu, Dacheng Tao, and Meng Wang.
\newblock Adaptive hypergraph learning and its application in image
  classification.
\newblock \emph{IEEE Transactions on Image Processing}, 21\penalty0
  (7):\penalty0 3262--3272, 2012.

\bibitem[Yu et~al.(2013)Yu, Tao, Rui, and Cheng]{Yu2013}
Jun Yu, Dacheng Tao, Yong Rui, and Jun Cheng.
\newblock Pairwise constraints based multiview features fusion for scene
  classification.
\newblock \emph{Pattern Recognition}, 46\penalty0 (2):\penalty0 483--496, 2013.

\bibitem[Yu et~al.(2015{\natexlab{a}})Yu, Hong, Tao, and Wang]{Yu2015}
Jun Yu, Chaoqun Hong, Dapeng Tao, and Meng Wang.
\newblock Semantic embedding for indoor scene recognition by weighted
  hypergraph learning.
\newblock \emph{Signal Processing}, 112:\penalty0 129--136, 2015{\natexlab{a}}.

\bibitem[Yu et~al.(2015{\natexlab{b}})Yu, Tao, Wang, and Rui]{Yu2015a}
Jun Yu, Dacheng Tao, Meng Wang, and Yong Rui.
\newblock Learning to rank using user clicks and visual features for image
  retrieval.
\newblock \emph{IEEE transactions on cybernetics}, 45\penalty0 (4):\penalty0
  767--779, 2015{\natexlab{b}}.

\bibitem[Yu et~al.(2017{\natexlab{a}})Yu, Yang, Gao, and Tao]{Yu2017a}
Jun Yu, Xiaokang Yang, Fei Gao, and Dacheng Tao.
\newblock Deep multimodal distance metric learning using click constraints for
  image ranking.
\newblock \emph{IEEE transactions on cybernetics}, 2017{\natexlab{a}}.
\newblock \doi{10.1109/TCYB.2016.2591583}.

\bibitem[Yu et~al.(2017{\natexlab{b}})Yu, Zhang, Kuang, Lin, and Fan]{Yu2017}
Jun Yu, Baopeng Zhang, Zhengzhong Kuang, Dan Lin, and Jianping Fan.
\newblock iprivacy: image privacy protection by identifying sensitive objects
  via deep multi-task learning.
\newblock \emph{IEEE Transactions on Information Forensics and Security},
  12\penalty0 (5):\penalty0 1005--1016, 2017{\natexlab{b}}.

\bibitem[Yuan et~al.(2015)Yuan, Mou, and Lu]{scenerecognition}
Yuan Yuan, Lichao Mou, and Xiaoqiang Lu.
\newblock Scene recognition by manifold regularized deep learning architecture.
\newblock \emph{IEEE transactions on neural networks and learning systems},
  26\penalty0 (10):\penalty0 2222--2233, 2015.

\bibitem[Zhou et~al.(2014)Zhou, Lapedriza, Xiao, Torralba, and
  Oliva]{places205}
Bolei Zhou, Agata Lapedriza, Jianxiong Xiao, Antonio Torralba, and Aude Oliva.
\newblock Learning deep features for scene recognition using places database.
\newblock In \emph{Advances in neural information processing systems}, pages
  487--495, 2014.

\end{thebibliography}

\end{document}